\documentclass[runningheads]{llncs}

\usepackage{eccv}

\usepackage{eccvabbrv}

\usepackage{graphicx}
\usepackage{booktabs}
\usepackage{siunitx}

\usepackage[accsupp]{axessibility}  %

\DeclareSIUnit{\FLOP}{FLOP}
\DeclareSIUnit{\SOP}{SOP}

\graphicspath{{figures/}}

\usepackage{hyperref}

\usepackage{orcidlink}

\makeatletter
\newcommand{\printfnsymbol}[1]{%
	\textsuperscript{\@fnsymbol{#1}}%
}
\makeatother

\begin{document}

\title{Drone Detection Using a Low-Power Neuromorphic Virtual Tripwire}

\author{Anton Eldeborg Lundin\thanks{equal contribution}\inst{1}\orcidlink{0009--0008--5842--1883}  \and
Rasmus Winzell \printfnsymbol{1}\inst{1}\orcidlink{0009--0003--6697--2902} \and
Hanna Hamrell \inst{1}\orcidlink{0009-0006-3557-5295} \and
David Gustafsson \inst{1}\orcidlink{0000--0002--4370--2286} \and
Hannes Ovrén \inst{1}\orcidlink{0000--0003--3365--4062}}
\authorrunning{A.~Eldeborg Lundin et al.}

\institute{Swedish Defence Research Agency (FOI), 583 30 Linköping, Sweden,
\email{\{anton.eldeborg.lundin,rasmus.winzell,hanna.hamrell,david.gustafsson, \\ hannes.ovren\}@foi.se},
\url{https://foi.se}}

\maketitle

\begin{abstract} 
  Small drones are an increasing threat to both military personnel and civilian infrastructure,
  making early and automated detection crucial.
  In this work we develop a system that uses spiking neural networks and neuromorphic cameras (event cameras) to detect drones.
  The detection model is deployed on a neuromorphic chip making this a fully neuromorphic system.
  Multiple detection units can be deployed to create a virtual tripwire which detects when and where drones enter a restricted zone.
  We show that our neuromorphic solution is several orders of magnitude more energy efficient
  than a reference solution deployed on an edge GPU, allowing the system to run for over a year on battery power.
  We investigate how synthetically generated data can be used for training,
  and show that our model most likely relies on the shape of the drone rather than the temporal characteristics of its propellers.
  The small size and low power consumption allows easy deployment in contested areas or locations that lack power infrastructure.

  \keywords{UAV \and Event camera \and Spiking neural network \and Neuromorphic}
\end{abstract}

\begin{figure}[tb]
  \centering
  \includegraphics[width=\linewidth]{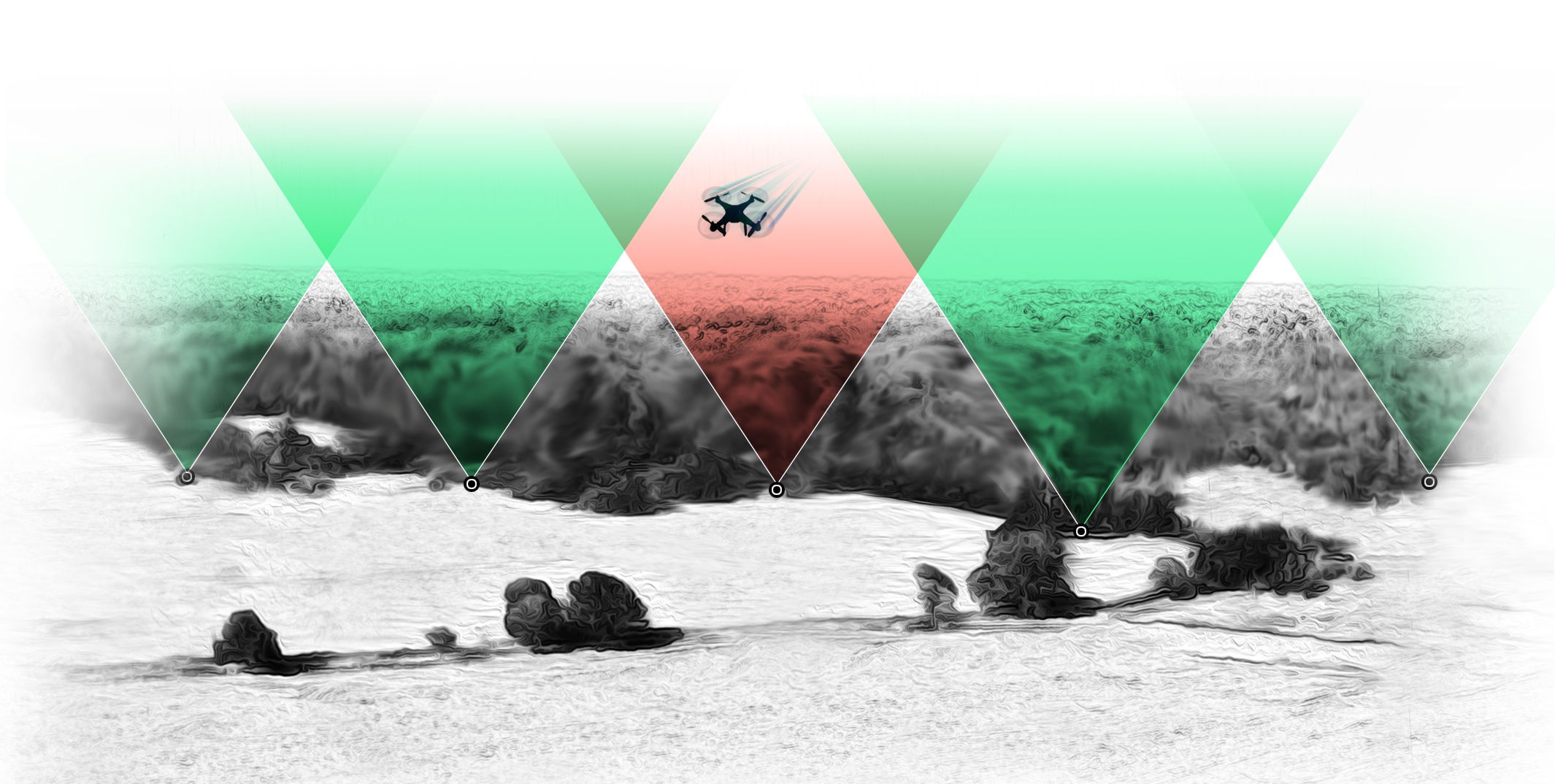}
  \caption{The system detects and gives an alarm whenever a drone enters the field of view of the event camera. Illustration by Esther Eriksson.}
  \label{fig:illustration}
\end{figure}

\section{Introduction} %
\label{sec:intro} %
In 2013 a small unmanned aerial vehicle (UAV), or \emph{drone}, landed on a stage next to the german chancellor and minister of defence, who where attending an election campaign rally \cite{heine2013}.
This was fortunately a benign political stunt, but it foreshadowed the present day where drones are increasingly being used as weapons.
In military conflicts drones are currently used for reconnaissance, as spotters for artillery, and as loitering munitions or cheap cruise missiles by strapping them with explosives.
This poses great threats to military personnel and equipment, as well as critical civilian infrastructure such as power plants or factories.
The detection of drones has therefore become an important task.

Several sensors and methods can be used to detect nearby drones. 
Radar-based systems can detect drones at large distances, but are expensive and easy to locate since they actively emit signals.
A simple and efficient solution that is also passive, is to use radio receivers to detect the radio link between the drone and its operator.
However, this will fail if the drones are programmed to autonomously navigate to and engage with the target without the need for a human operator,
which makes them impervious to jamming of the radio link.
There are strong indications that such autonomous drones have already been used in combat \cite{adam2024,mozur2024}.

In this work we describe a system for drone detection that uses vision to trigger an alarm whenever a drone enters the field of view of the sensor.
Placing several such sensor units next to to each other creates a virtual tripwire that can warn about an incoming drone and provide the location where it entered the restricted zone, see \cref{fig:illustration}.
It is unlikely that the location where the units are deployed has available electric power infrastructure, and the units must therefore be powered by battery, or harvest their own power (\eg using solar panels).
Since the deployment location could be difficult to reach, or even be contested territory, it is important that the units can operate for a long time before they run out of power.
At the same time it is preferable to avoid large batteries since this makes deployment and concealment more difficult.
This makes power usage an important design parameter.

In this work we use the combination of a neuromorphic vision sensor (the \emph{event camera}), spiking neural networks (SNN), and a neuromorphic computing device
to build a drone detection system that uses very little power.
To the best of our knowledge this is the first example of a fully neuromorphic system for drone detection.
The neuromorphic solution is evaluated against a more traditional solution using non-spiking artificial neural networks (ANN) deployed on a GPU,
in terms of both detection performance and power consumption.
We further describe how we use machine learning to train an SNN-based detection model that is deployed on a commercially available neuromorphic device,
and give insights into the possibility of using synthetically generated training data.

\section{Related Work} %
Here we focus on previous works that use event cameras for drone detection.
An overview of alternative solutions for drone detections can be found in \cite{wang2021}.

Our approach is inspired by the works of Stewart \etal \cite{stewart2021,stewart2022} who similarly use upwards facing event cameras to detect the presence of drones within the sensor field of view.
Their approach exploits the fact that drone propellers typically generate frequencies in the event data that are not common for other airborne objects, such as birds.
Histograms of times between ON-OFF-ON (and OFF-ON-OFF) events are used to represent the frequency content of the scene, and in their earlier work
they apply a classifier based on either the FFT or curve fitting to these histograms \cite{stewart2021}.
In their later work they use a higher resolution event camera and a small neural network for classification which increases the detection range from \qty{9}{\meter} to \qty{19}{\meter}
and the inference speed from \qty{1}{\Hz} to \qty{100}{\Hz} \cite{stewart2022}.
Their system is based on a Raspberry Pi 4 with a total power consumption of approximately \qty{5}{\watt}.

Sanket \etal \cite{sanket2021} also exploit the distinctive frequency of propellers to detect drones.
They use convolutional neural networks that take event frames as input, and train on synthetically generated data.
Their system runs at \qty{35}{\Hz} on a Google Coral edge TPU (\qty{17}{\watt}).

Kirkland \etal \cite{kirkland2019} use a convolutional SNN trained using spike-time dependent plasticity (STDP) to detect drones.
The detection is mostly based on visual appearance and does not explicitly model temporal information generated by the propellers.
As input they use aggregated event frames with varying integration times, and they do not report any estimates of power consumption.

The Speck neurmorphic chip \cite{yao2024} that is used in this work was only recently made available, and there are few published works that use it.
Relevant to us is the work by Caccavella \etal \cite{caccavella2023} who used the Speck for face detection and provide valuable insights
into feasible network architectures and training strategies.
The final layer in their network that peforms bounding box regression was however not performed on the neuromorphic chip, but on a host CPU.

\section{Method}
The full system consists of several identical sensor units that are deployed with an overlapping field of view, see \cref{fig:illustration}.
When a drone enters the field of view of one unit it sends an alarm via \eg a radio link to let the user know when and where an intrusion occured.
Here we focus on how a single unit can be designed for low power drone detection, and do not consider how this detection is transmitted to and handled
by some external system.

\subsection{Overview} %
\label{sec:system_overview}
We base our system on the Speck neuromorphic device from Synsense \cite{yao2024,synsense-speck-dev-kit-datasheet}.
The Speck consists of an $128 \times 128$ pixel event camera that is directly connected to a neuromorphic chip (DYNAP-CNN)
which runs a user-defined convolutional spiking neural network.
By current standards the size of the networks that can be deployed on the Speck are quite small,
and is limited to nine layers with a total of 328k neurons.
Each layer performs asynchronous convolution, integrate-and-fire spiking activation, and an optional sum pooling.

In contrast to traditional cameras an event camera does not provide images at set times.
Instead, a sensor element at position $(x, y)$ generates an \emph{event}, $e = (x, y, t, p)$, at any time, $t$, when the difference in (logarithmic) light intensity crosses a threshold.
The polarity, $p = \{+1,-1\}$, indicates whether the light intensity has decreased or increased.
Event cameras have several benefits over traditional cameras, such as a high temporal resolution, high dynamic range, and low power consumption.
For a more thourough overview of event cameras, see \cite{gallego2022}.

The system is deployed with the event camera on the Speck facing upwards
using the \qty{3.62}{\mm} lens that was supplied with the Speck Development Kit.
The events from the event camera are handled asynchronously, event-by-event, by the neuromorphic chip
where the deployed SNN performs binary classification to decide whether a drone is in view or not.

\subsection{Model Training}
\label{sec:model_training}
\begin{figure}[tb]
	\centering
	\includegraphics[width=0.7\textwidth]{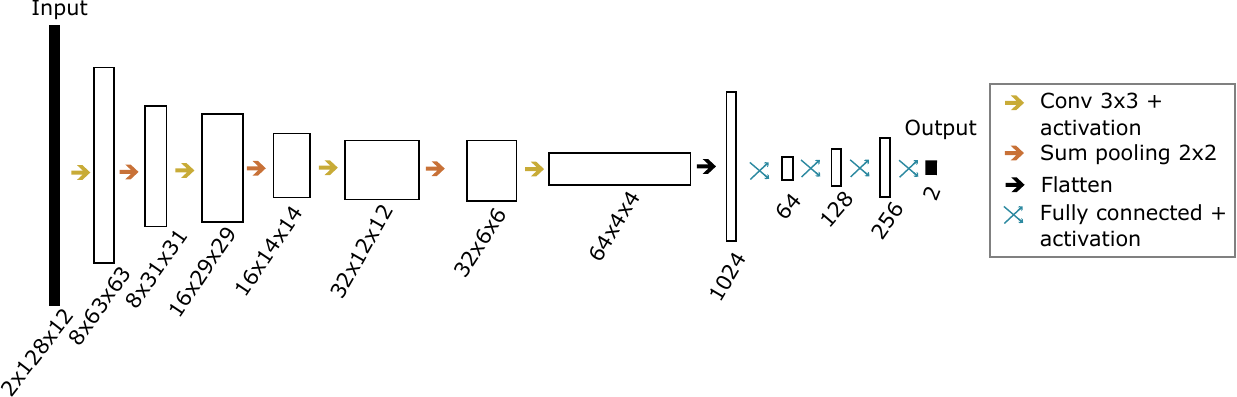}
	\caption{The architecture of the SNN and the ANN used in the experiments. The black rectangle represents the input data during training.}
	\label{fig:architecture}
\end{figure}

To allow performance comparisons we created two different neural network models: one spiking neural network (SNN) and one equivalent artificial neural network (ANN).
The two networks share the same architecture (see \cref{fig:architecture}) which is loosely inspired by \cite{caccavella2023} and is designed to adhere to the limits set by the Speck in terms of number of (hardware) layers and size of the weights.
The input is defined by the resolution of the Speck sensor ($128\times 128$ pixels) with one channel each for positive and negative event polarity. 
The first four layers are 2D convolutions followed by max pooling (for the first three). This is then followed by a flattening of the feature vector and four fully connected layers.
Since the goal is binary classification of the scene, there are two output classes - \textit{drone} and \textit{no-drone} -
and consequently there are two output neurons, one for each class.

It is recommended that models deployed on the Speck sets the bias of all layers to zero to save on computational operations and thus power \cite{synsense-dynapcnndoc},
and we apply this to both the SNN and ANN model.
Sum pooling is used for all pooling operations since that is what is implemented on the Speck.

The drone detection performance is evaluated using recall and false discovery rate (FDR).
We aim for high recall, to maximize the probability of detecting drones, while keeping a low FDR to not give unneccessary false alarms.

\subsubsection{Spiking Neural Network} %
The Speck does not support on-chip training, and therefore training was performed off-chip on a regular desktop GPU using Sinabs\cite{synsense-sinabs}.
We train our SNN using standard machine learning methods \cite{eshraghian2023} using backpropagation through time (BPTT) and surrogate gradients \cite{neftci2019}.
While the Speck supports the leaky integrate and fire (LIF) spiking neuron model, it is not recommended since it requires an external synchronous clock and use substantially more power \cite{synsense-dynapcnndoc}.
For this reason we use the integrate and fire (IF) spiking neuron model for all layers in the network.
Informed by \cite{caccavella2023,synsense-dynapcnndoc} we use the \emph{MultiSpike} spike generation function in Sinabs instead of \emph{SingleSpike}, since this best matches the hardware.

We split the data sequences into shorter \qty{50}{\milli\second} samples, and the asynchronous arrival of events is simulated by converting the events to event frames based on small time steps. 
The sample length was chosen to capture a few turns of the drone propellers within a single sample (turning with about 9000 rpm), 
and a time step of \qty{1}{\milli\second} was used to strike a balance between computational time and temporal resolution.

As loss function we applied a mean squared error (MSE) loss at each time step with a target spike count of one and zero for the correct and incorrect class, respectively.
For training we used the Periodic Exponential\cite{synsense-sinabs} surrogate gradient function, and the Adam optimizer.

Regularization of the synaptic operations can be used to produce more energy efficient networks \cite{caccavella2023},
and we investigate this possibility using two different regularization loss functions.
We use a synaptic operations regularization

\begin{equation}
	L_\text{SOP} = \alpha(S_0-\sum_l s^l)^2,
\end{equation}
where $S_0$ is the target synaptic operations for all layers of the network and $s^l$ is the number of synaptic operations for layer $l$ \cite{sorbaro2020optimizing}.
We set $\alpha=\frac{10}{S_0^2}$ after performing parameter search on the nominator.

In addition we use a weight regularization inspired by \cite{ziegler2024spiking}, defined as
\begin{equation}
	L_\text{weight} = \sum_l  \max_{k}|W_k^l|,
\end{equation}
where $W_k^l$ is a single weight parameter of layer $l$, to penalize the magnitude of the weights.

Combining the target MSE loss function, $L_\text{MSE}$, and the regularizations, the final loss function is defined as
\begin{equation}
	L = L_\text{MSE} + L_\text{SOP} + L_\text{weight}.
\end{equation}

\subsubsection{Non-spiking Equivalent Network}
\label{sec:ann_equivalent}
The equivalent ANN was implemented and trained using PyTorch with cross entropy loss without any regularization.
The IF spiking neuron model used for each layer in the SNN was replaced by ReLU activation functions.

Instead of the multiple time steps used for training the SNN the event data was converted to single channel frames by aggregating all events in the \qty{50}{\milli\second} time window.

\subsection{Dataset} %
\label{sec:dataset}
A larger dataset with real data was used for training the SNN and ANN, and a smaller synthetic dataset was used to test the generalizability of the models.

\subsubsection{Speck UAV Dataset}
\begin{figure}
	\centering
	\begin{subfigure}{.49\textwidth}
		\centering
		\includegraphics[width=.4125\textwidth]{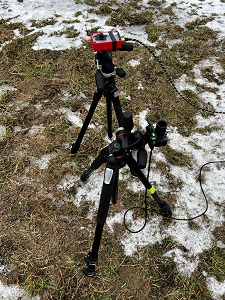}
		\includegraphics[width=.55\textwidth]{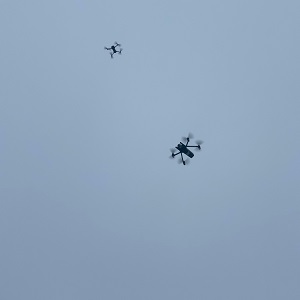}
		\caption{Field}
		\label{fig:field_setup}
	\end{subfigure}
	\begin{subfigure}{.49\textwidth}
		\centering
		\includegraphics[width=.4125\textwidth]{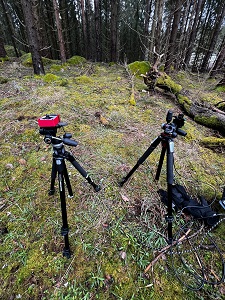}
		\includegraphics[width=.55\textwidth]{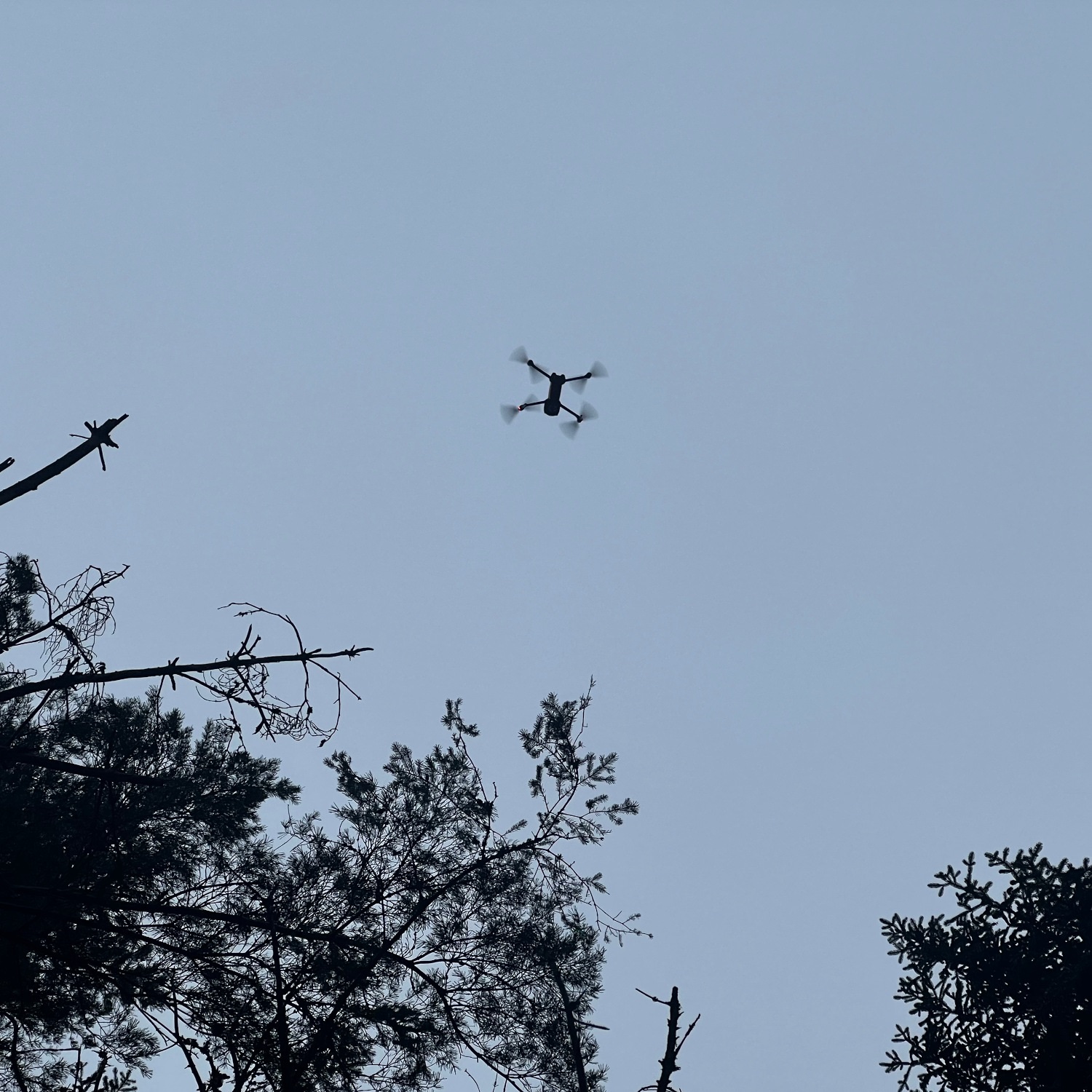}
		\caption{Forest meadow}
		\label{fig:meadow_setup}
	\end{subfigure}
	\caption{Sensor setup (left) and sample visual image (right) at both locations. The Speck Development Kit is inside the red box cover.}
	\label{fig:sensor_setup}
\end{figure}

Data was collected using a Speck development kit during a single day in February 2024 in southern Sweden.
The sequences were recorded in two different locations, in an open field and a forest meadow, 
where the main difference between the two locations is the visibility of tree tops in the latter.
The weather during the recordings was overcast, creating a uniform background.
\Cref{fig:sensor_setup} shows the sensor setup and example visual images from the two locations.

\begin{figure}
	\centering
	\begin{subfigure}{.3\textwidth}
		\centering
		\includegraphics[width=.75\linewidth]{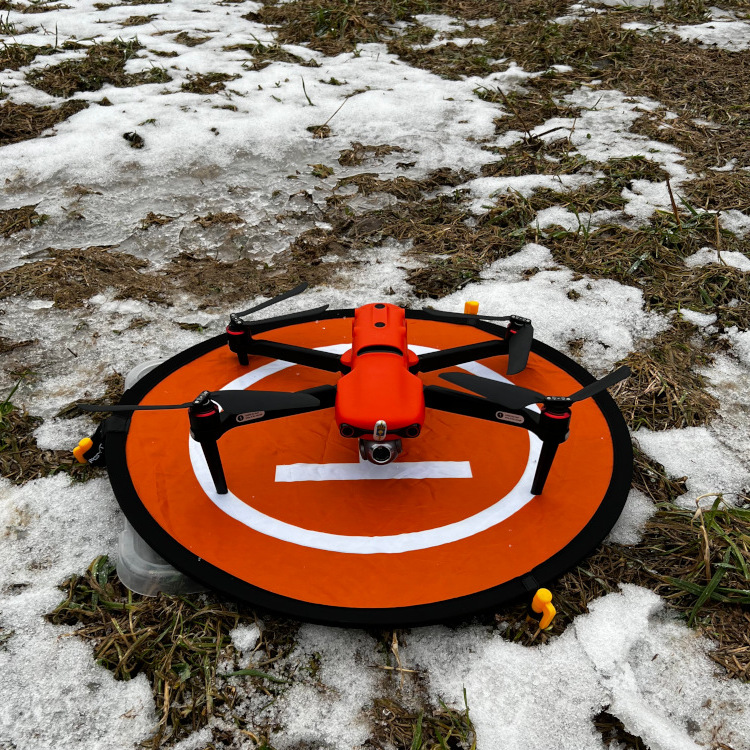}
		\caption{Autel Evo}
		\label{fig:autelevo}
	\end{subfigure}
	\begin{subfigure}{.3\textwidth}
		\centering
			\includegraphics[width=.75\linewidth]{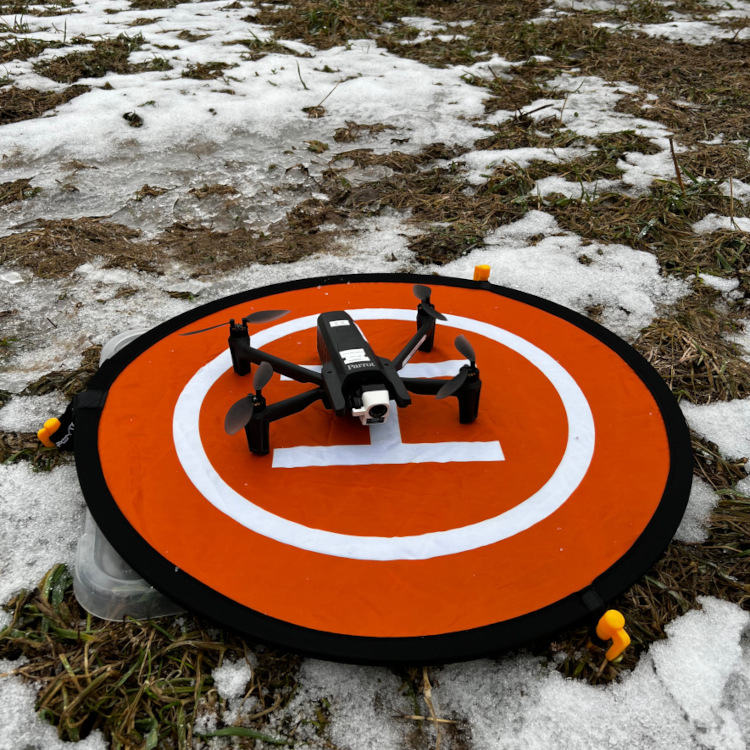}
		\caption{Parrot}
		\label{fig:parrot}
	\end{subfigure}
	\begin{subfigure}{.3\textwidth}
		\centering
		\includegraphics[width=.75\linewidth]{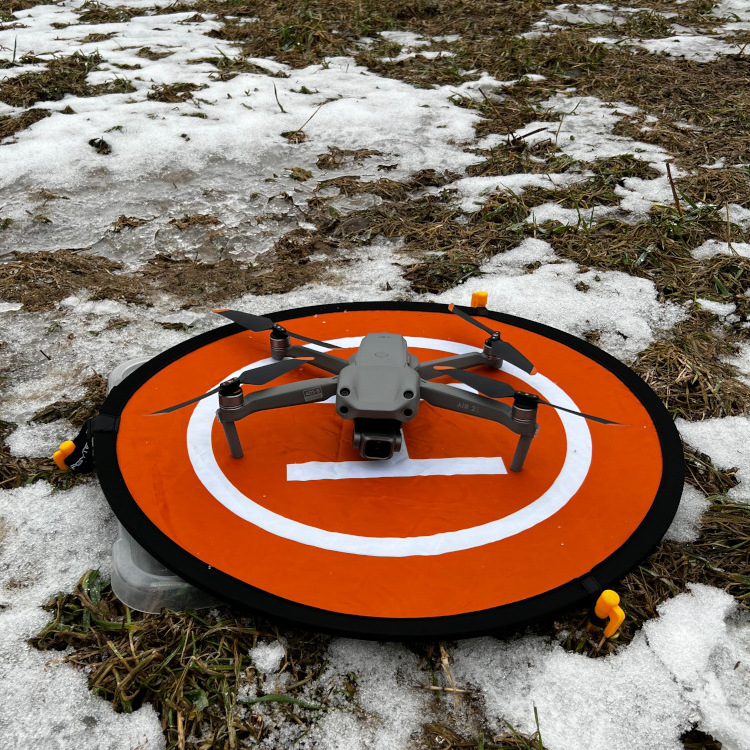}
		\caption{DJI Air}
		\label{fig:djiair}
	\end{subfigure}
	\caption{The drones used for the collection of data.}
	\label{fig:drones}
\end{figure}

Three commercially available drones were used as targets: Parrot Anafi, DJI Air, and Autel Evo 2, see \cref{fig:drones}.
The drones were flown at various altitudes and velocities with different types of movement.
The targeted altitudes where \qtylist{5;10;15;20;25}{\meter} and the movements were categorized as 
\textit{flying straight slow}, \textit{flying straight fast}, \textit{lateral movement}, or \textit{rotating}.
The relatively low altitudes were chosen to match the optics used in the development kit, 
however, these optics can easily be changed to allow more relevant altitudes in future work.
Most sequences contain a single drone, but a few sequences with multiple drones were also recorded, as well as one sequence where a drone is ascending rapidly.
Negative (non-drone) sequences include recordings with only moving tree branches and balls that were thrown in the event camera field of view.
Two more difficult positive sequences contains two drones flying at different heights, where one of the sequences additionally has a ball thrown below the drones.
All time spans of the recorded data where at least a part of a drone is visible was labeled as \textit{drone}, and all other as \textit{no-drone}.

\subsubsection{Synthetically Generated Data} %
\label{sec:synthetic_data}
Synthetic data can be used to expand the training dataset and simplify data annotation, and is especially useful when it is impractical to record real data.
We generate synthetic event camera data by rendering high frame-rate video in Blender which is then converted to an event data stream using a method based on \cite{gehrig2020}.

\begin{figure}
	\centering
	\begin{subfigure}{.3\textwidth}
		\centering
		\includegraphics[width=.8\linewidth,angle=-90]{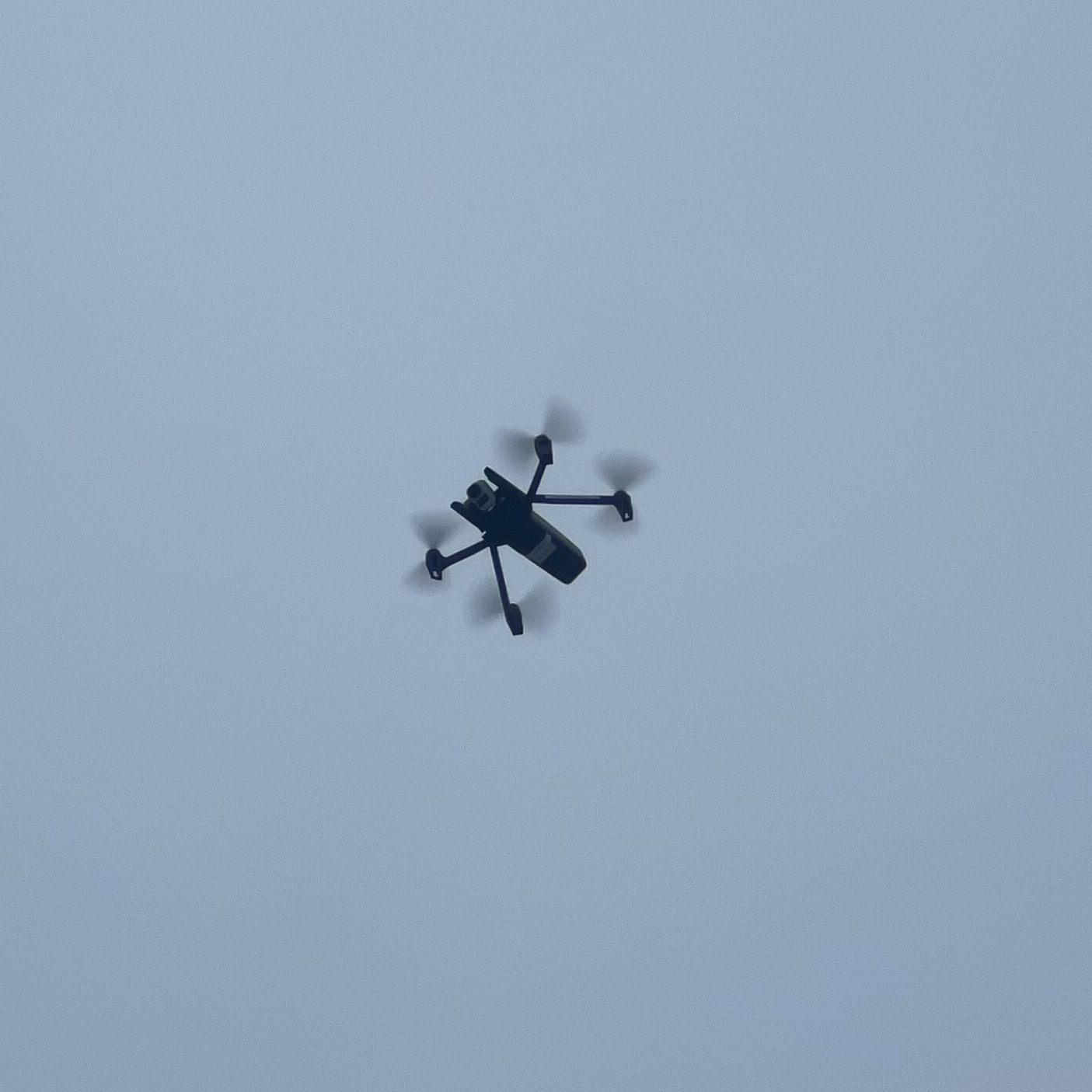}
		\caption{Real}
		\label{fig:real_drone}
	\end{subfigure}
	\begin{subfigure}{.3\textwidth}
		\centering
		\includegraphics[width=.8\linewidth,angle=-90]{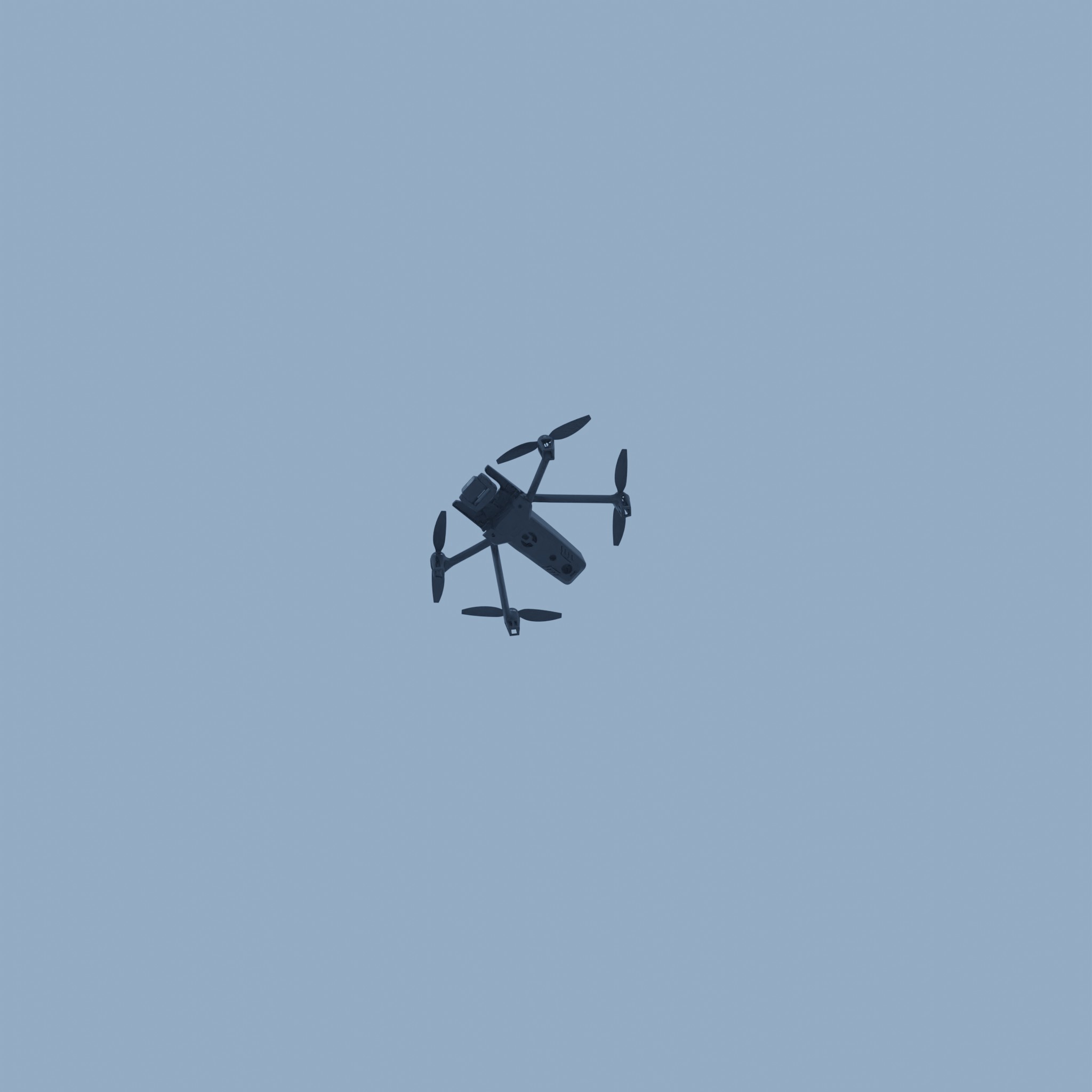}
		\caption{Simulated}
		\label{fig:synthetic_drones}
	\end{subfigure}
	\caption{Comparison between a real and simulated DJI Air drone.}
	\label{fig:real_vs_synthetic_drone}
\end{figure}

Our drone 3D model consists of a combination of the body of a Parrot Anafi and the propellers and camera system of a DJI Mavic Mini.
This combination was motivated by availability of 3D models and a similar visual appearance to the real data, see \cref{fig:real_vs_synthetic_drone}.
The initial position and orientation of the drone was randomized in the sensor field of view and the propeller angular velocity was randomly set to approximately match the real drones used in the Speck UAV dataset. %
One test set was also generated where the drone propellers were completely removed to investigate whether the trained models
learnt the temporal characteristics of the propellers, or only the shape of the drone.

To properly simulate event generation from video the maximum optical flow between two subsequent frames must be below one pixel.
This implies a constraint on the image plane velocity of the tip of the drone propeller,
\begin{equation}
	v_\text{tip}=\frac{\pi d_\text{prop} f_\text{prop}}{f_\text{video}} \leq 1,
\end{equation}
where $f_\text{video}$ is the video frame rate, $d_\text{prop}$ is the diameter of the propeller in pixels, and $f_\text{prop}$ its rotational frequency.
The minimum frame rate is therefore given by

\begin{equation}
	f_\text{video} \geq \pi d_\text{prop} f_\text{prop}.
\end{equation}
With the used drone model and rotational frequencies observed in the real data, the frame rate of the rendering was set to \qty{10000}{fps}.

In \cite{gehrig2020} an event is generated when 
\begin{align}
	\Delta L \geq p C \,,
\end{align}
where $\Delta L$ is the difference in log intensity, $p$ is event polarity, and $C$ a threshold.
Due to the high contrast in the data, and inspired by the \emph{MultiSpike} spike generation function (see \cref{sec:model_training}),
our implementation is extended to instead generate
\begin{equation}
	N_e = \left\lfloor \frac{{\Delta L} p}{C} \right\rfloor
\end{equation}
events when the threshold is passed.

\subsubsection{Partitioning of the Data}
The collected data was split into training, validation and test data.
All sequences that contained the DJI Air drone were reserved for testing to evaluate how the model generalized to an unseen drone model.
\Cref{tab:test_sequences} describes the test set,
and the remaining data was split into $95\%$ training data and $5\%$ validation data, where the small set of validation data was motivated by the limited amount of data. The synthetically generated data was only used to evaluate the generalizability of the model and was not used for training.

\begin{table}[tb]
	\caption{Test data sequences.}
	\label{tab:test_sequences}
	\centering
	\begin{tabular}{@{}p{0.3\linewidth} p{0.7\linewidth}@{}} %
		\toprule
		\textbf{Sequence name} & \textbf{Description}\\
		\midrule
		DJI Air 5 m  & \\
		DJI Air 10 m &  \\
		DJI Air 15 m & The DJI Air drone flies at the specified altitude.\\ %
		DJI Air 20 m & \\
		DJI Air 25 m & \\
		\midrule
		Parrot  & Parrot Anafi where altitude varied during flight at the open field location.\\
		\midrule
		Two Drones & One DJI Air and one Parrot Anafi drone flies at an altitude of 20 and 10 meters, respectively, at the open field location.\\
		\midrule
		Two Drones and Ball & Same as Two Drones, except a ball is thrown between the sensor and the drones.\\
		\midrule
		Forest & The Autel Evo 2 drone flies at various altitudes in the forest.\\
		\bottomrule
	\end{tabular}
\end{table}

\section{Experiments}
\label{sec:experiments}
We first investigate how different altitudes in the training data affect the performance of the trained SNN,
and use this to select the model that is then used for the other experiments.

\subsection{Training of the SNN}
\label{sec:exp_snn_training}

\begin{figure}[tb]
	\centering	
	\begin{subfigure}{.49\textwidth}
		\centering
		\includegraphics{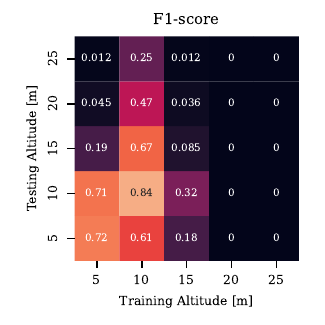}
		\caption{Without regularization}
		\label{fig:score_matrix_no_reg}
	\end{subfigure}
	\begin{subfigure}{.49\textwidth}
		\centering
		\includegraphics{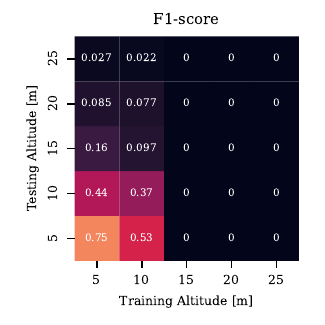}
		\caption{With regularization}
		\label{fig:score_matrix_reg}
	\end{subfigure}
	\caption{F1-score for models trained with and without regularization and evaluated with different altitude splits of the training data.}
	\label{fig:score_matrix}
\end{figure}

\begin{figure}[tb]
	\centering	
	\includegraphics{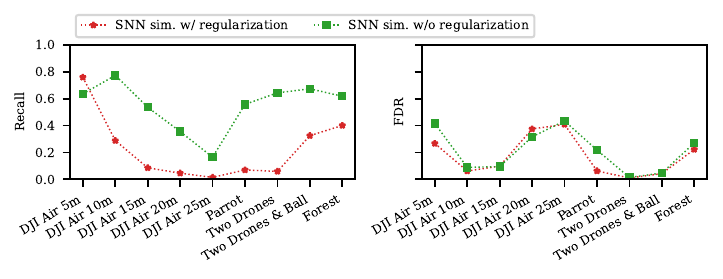}
	\caption{Recall (left) and FDR (right) for networks trained with regularization on \qty{5}{\meter} data, and without regularization on \qty{10}{\meter} data, respectively.}
	\label{fig:regularization_eval}
\end{figure}

Training an SNN on the full training dataset, with all drone altitudes, proved to be challenging.
Instead, models were trained on each altitude separately, and were then evaluated against all test sequence altitudes to find the model that generalized the best.
In addition, each model was trained with and without regularization (see \cref{sec:model_training}).
\Cref{fig:score_matrix} shows that the models trained on 5 and 10 meters data, respectively, achieved the best F1 scores at their own altitude, while also performing decently at the other altitudes.
The models trained on data from 15, 20, or 25 meters performed poorly at all test altitudes, including the training altitude, likely due to the limited information in those sequences.

Comparing \cref{fig:score_matrix_no_reg} and \cref{fig:score_matrix_reg} we see that the model trained with and without regularization performed the best on \qty{5}{\meter} and \qty{10}{\meter} data, respectively.
In \cref{fig:regularization_eval}, the recall and FDR for both models is presented, showing that they perform well at and close to the altitudes they were trained on.

Since the network trained without regularization on \qty{10}{\meter} data performed best in simulation, it was selected for deployment on the Speck.

\subsection{Classification}
In \cref{sec:exp_snn_training} we found the SNN that performed best in simulation with Sinabs.
Since the simulation is not expected to be perfect, the final evaluation of classification performance must be
done with the network deployed on the hardware.

The simulated and deployed SNNs are then compared to the equivalent ANN described in \cref{sec:ann_equivalent}.
Note that the ANN was trained on the full dataset with all altitudes, and not only on a single altitude as was the case for the selected SNN.

\begin{figure}[tb]
	\includegraphics{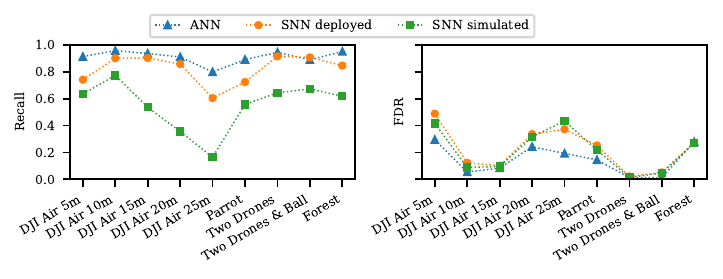}
	\caption{Recall (left) and FDR (right) for the evaluated networks.}
	\label{fig:class_eval}
\end{figure}

The three networks (simulated SNN, SNN deployed on Speck, and ANN) are evaluated with respect to the FDR and recall on the full test set.
\Cref{fig:class_eval} shows that the ANN performs well on all altitudes, while the SNN in general performs worse on higher altitudes, farther from the \qty{10}{\meter} on which it was trained.

Interestingly, the performance of the SNN, in terms of recall, is considerably higher when deployed on hardware compared to the simulation in Sinabs.
Since the FDR is similar to the simulated SNN, this is not a case of the deployed network simply classifying all samples as drones.
The cause for the difference in performance between the deployed and simulated networks was not investigated further.

\subsection{Synthetic vs Real Data}

\begin{figure}[tb]
	\includegraphics{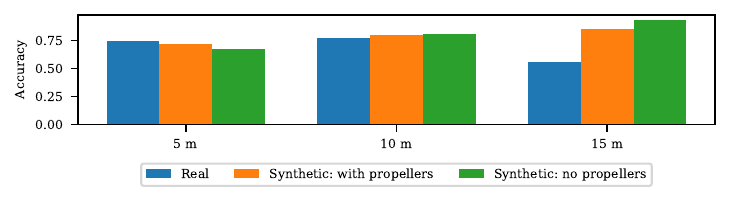}
	\caption{Accuracy of the simulated network on synthetically generated and real data, for different heights.}
	\label{fig:synthetic_acc}
\end{figure}

Using the synthetically generated data from \cref{sec:synthetic_data} we want to investigate two things.
How well does the model trained on real data generalize to synthetic data, and did the SNN learn the temporal characteristics of the propellers, or only the shape of the drone? We generated two sets of synthetic data: one with spinning propellers, and one where the propellers were entirely removed. The data was generated for altitudes of \qtylist{5;10;15}{\meter} since that is the range where our SNN performs the best. 

If the network has learned the temporal characteristics of the drone propellers we would expect a lower accuracy on the test data without propellers.
However, \cref{fig:synthetic_acc} shows that this is not the case, and we can therefore be quite certain that the SNN makes its decision
based on the shape of the drone body. Further, the domain gap between synthetic and real data is small for \qtylist{5;10}{\meter}, while there is a larger gap at \qty{15}{\meter}. This indicates that using synthetic data can be viable under certain conditions.

\subsection{Power Consumption} %
We compare the power consumption between the SNN deployed on the Speck
with an ANN deployed on a consumer GPU for edge applications.

We model the total power consumption as
\begin{align}
  P_\text{total} = P_\text{idle} + P_\text{dynamic},
\end{align}
where $P_\text{idle}$ is the constant power draw when no computations are performed
and $P_\text{dynamic}$ is the added power needed to perform the computations.
The latter is not constant but will rather depend on \eg the size of the deployed network.

\subsubsection{Speck Power Consumption}
The power consumption of the Speck depends on the number of synaptic operations (SOP),
which in turn depends on both the network architecture, the network weights, and the input.
To estimate a model of the power consumption we initialized the network with random weights and
ran inference on the full test set while recording the power consumption.
The power consumption was measured by the diagnostic tools available on the Speck development kit.
This was repeated multiple times with increasing average weight values to force the network to
produce more spikes, and thus more synaptic operations.
From the measurements we fit an affine model to compute the total power of the Speck
\begin{align}
  P^\text{Speck} = P_\text{idle}^\text{Speck} + k N_\text{SOP},
  \label{eq:speck_energy_model}
\end{align}
given the number of synaptic operations, $N_\text{SOP}$, and constants
$P_\text{idle}^\text{Speck}= \qty{1.48}{\mW}$ and $k = \qty{11.53d-6}{\milli\watt\per\SOP}$.

\begin{figure}[tb]
  \centering
  \includegraphics{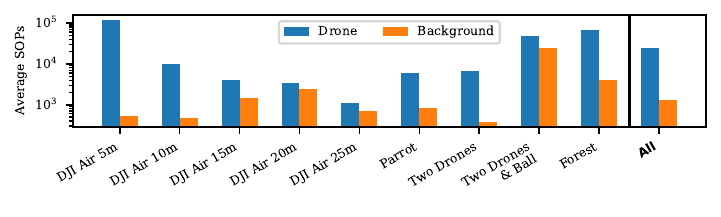}
  \caption{Number of synaptic operations for drone and no-drone samples in the test set.}
  \label{fig:testset_sops}
\end{figure}

Using the trained and deployed network from \cref{sec:exp_snn_training} we measure the number of synaptic operations for both \emph{drone} and \emph{no-drone} samples in the test set, see \cref{fig:testset_sops}. This results in a dynamic power for our use case that varies between \qtyrange[range-phrase = --]{0.3}{5.65}{\mW} depending on whether a drone is visible (highest power) or not (lowest power). The estimated static power, \qty{1.48}{\mW}, is added to this.

\subsubsection{ANN Power Consumption}
The power consumption of the ANN depends on the network itself and the hardware on which it is deployed.
We use an Nvidia TX1 as our reference platform since it is intended for AI applications on the edge.
Also, the TX1 was previously used by Blouw \etal \cite{blouw2019} for a similar comparison between ANNs and SNNs, and they measured its idle power consumption to be $P_\text{idle}^\text{TX1} = \qty{2.64}{\watt}$.

The dynamic power consumption typically depends on the number of floating point operations (FLOP), $N_\text{FLOP}$, that the network performs during inference.
We adapt the method in \cite{desislavov2023} to compute the dynamic power of the TX1 as
\begin{align}
  P_\text{dynamic}^\text{TX1} &= \left ( \frac{P_\text{TDP} - P_\text{idle}^\text{TX1}}{T_\text{max}} \right ) N_\text{FLOP}.
\end{align}
The specifications for the TX1 states a thermal design power (TDP) of $P_\text{TDP} = \qty{6}{\watt}$, and maximum throughput (fp32) of $T_\text{max} = \qty{511}{\giga\FLOP\per\second}$.
The number of floating point operations performed by our ANN, for one sample, is $N_\text{FLOP} = \qty{5.62}{\mega\FLOP}$,
computed using the \texttt{ptflops} tool \cite{sovrasov2018}.
This yields a final dynamic power $P_\text{dynamic}^\text{TX1} = \qty{0.74}{\mW}$ for the TX1 and a total power consumption of \qty{2640.74}{\mW}.

\subsubsection{Operational Scenario}
\begin{figure}[tb]
  \centering
  \includegraphics{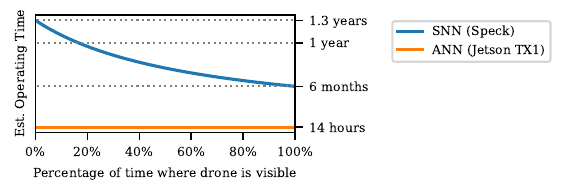}
  \caption{Operating time vs the time a drone is in view for a scenario where the system is powered by a \qty{37}{\watt\hour} battery.}
  \label{fig:power_consumption_scenario}
\end{figure}

To put the power consumption in context we assume that we want to power the system using a \qty{37}{\watt\hour} battery
with a self-discharge rate of 3\% per month.
The question is how long the system can operate using either an SNN or ANN solution,
for a given rate of drone observations.
The power consumption of the ANN deployed on the TX1, in contrast to the SNN on the Speck, does not depend on the rate of drone observations.
This is because detection needs to be performed for every frame, and the computations are the same
regardless of the input.
\Cref{fig:power_consumption_scenario} shows that the estimated operating time for the ANN on GPU solution is approximately 14 hours,
while the SNN on Speck solution operating time spans between \qty{6}{months} and \qty{1.3}{years} depending on the input. In practice, the system will rarely observe drones or other moving objects, likely resulting in an operating time of a year or more.

\subsection{Qualitative Results}

\begin{figure}[tbp]
	\centering		
	\begin{subfigure}{0.495\textwidth}
		\centering
		\includegraphics[width=\textwidth]{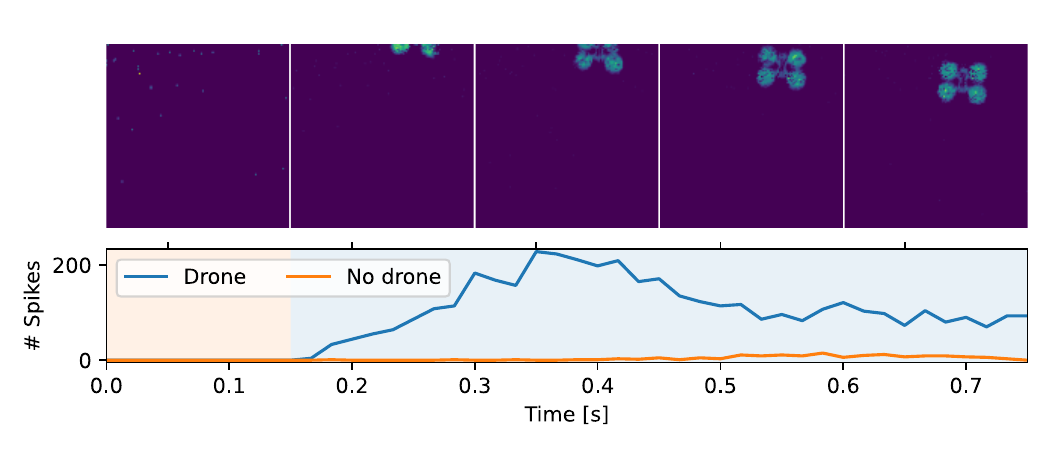}
		\caption{Horizontal transition}
		\label{fig:qualitative_transition}
	\end{subfigure}
	\begin{subfigure}{0.495\textwidth}
		\centering
		\includegraphics[width=\textwidth]{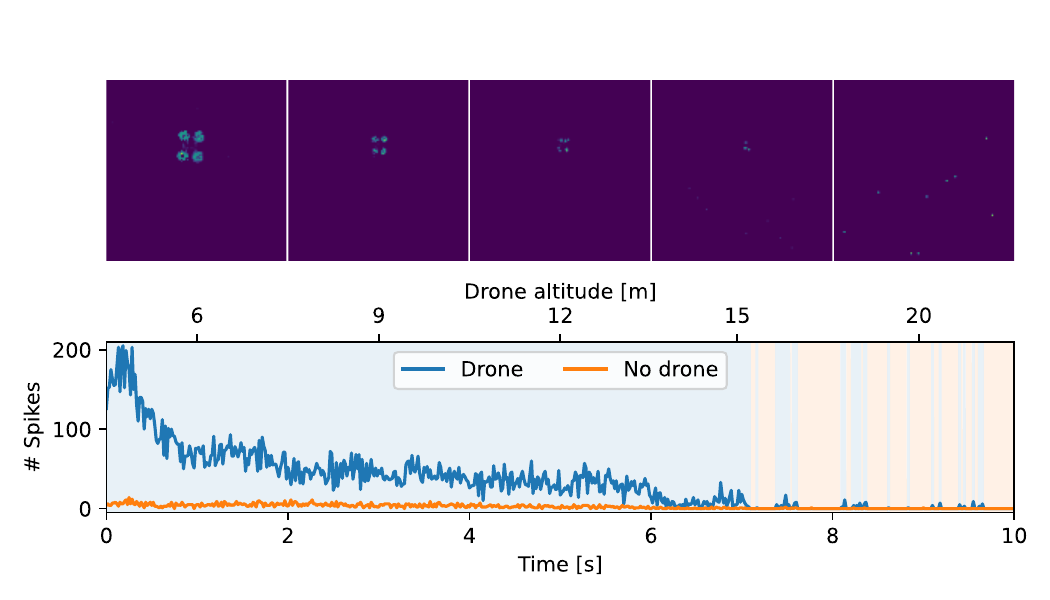}
		\caption{Drone moving upwards}
		\label{fig:qualitative_heights}
	\end{subfigure}
	\caption{Qualitative examples of test sequences evaluated using the SNN deployed on the Speck. 
	Top: example event frames. Bottom: Number of output spikes over time.
	The background color shows which class is the final system output (highest spike rate) for a given time span.}
	\label{fig:qualitative}
\end{figure}

\Cref{fig:qualitative} shows two qualitative examples using the SNN deployed on the Speck.
In \cref{fig:qualitative_transition} we can observe how the output spike rate of the drone class output neuron increases rapidly as the drone enters the field of view and that the correct class is given as ouput at all times.

\Cref{fig:qualitative_heights} shows another scenario where the drone is initially at an altitude of approximately \qty{5}{\meter}, and then ascends to an altitude above \qty{20}{\meter}.
The correct class is consistently predicted up to an altitude of around \qty{15}{\meter}, after which performance deteriorates and the correct class is only given sporadically.
This is consistent with the results previously shown in \cref{fig:class_eval}.

\section{Conclusion} %
We have described a system for drone detection that use both a neuromorphic sensor and compute device
that has a power consumption in the order of \qty{10}{\mW}.
The proposed neuromorphic system is several orders of magnitude more power efficient than the
ANN on GPU solution that was used for evaluation in this work (\qty{2.64}{\watt}),
as well as previous works (\qty{5}{\watt}\cite{stewart2021,stewart2022} and \qty{17}{\watt}\cite{sanket2021}).
Making fair comparisons of power consumption is hard, and while our reported power consumption
accounts for both sensor and computations, it does not account for any extra hardware required to \eg generate and send an alarm.

Our evaluation on synthetic data indicates that our SNN did not learn to detect drones based on their temporal characteristics,
but rather on their shape.
This could be the reason for the poorer performance on higher altitudes where the drone only occupies a few pixels.
It is possible that switching from an IF to LIF neuron model could solve this issue and allow detection also for higher altitudes
where there is little shape information.

Our results indicate that training an SNN is more difficult than an ANN, and that ANNs perform better with regard to detection performance.
However, we believe that the small decrease in detection performance is outweighed by the large increase in energy efficiency.

\bibliographystyle{splncs04}
\bibliography{main}

\begin{thebibliography}{10}
\providecommand{\url}[1]{\texttt{#1}}
\providecommand{\urlprefix}{URL }
\providecommand{\doi}[1]{https://doi.org/#1}

\bibitem{adam2024}
Adam, D.: Lethal {AI} weapons are here: how can we control them? Nature  \textbf{629},  521--523 (2024)

\bibitem{blouw2019}
Blouw, P., Choo, X., Hunsberger, E., Eliasmith, C.: Benchmarking keyword spotting efficiency on neuromorphic hardware. In: Proceedings of the 7th annual neuro-inspired computational elements workshop. pp.~1--8 (2019)

\bibitem{caccavella2023}
Caccavella, C., Paredes-Vallés, F., Cannici, M., Khacef, L.: Low-power event-based face detection with asynchronous neuromorphic hardware (2023), \url{https://arxiv.org/abs/2312.14261}

\bibitem{desislavov2023}
Desislavov, R., Mart{\'\i}nez-Plumed, F., Hern{\'a}ndez-Orallo, J.: Trends in ai inference energy consumption: Beyond the performance-vs-parameter laws of deep learning. Sustainable Computing: Informatics and Systems  \textbf{38},  100857 (2023)

\bibitem{eshraghian2023}
Eshraghian, J.K., Ward, M., Neftci, E.O., Wang, X., Lenz, G., Dwivedi, G., Bennamoun, M., Jeong, D.S., Lu, W.D.: {Training spiking neural networks using lessons from deep learning}. Proceedings of the IEEE. Institute of Electrical and Electronics Engineers  \textbf{111},  1016--1054 (2023). \doi{10.1109/jproc.2023.3308088}

\bibitem{gallego2022}
Gallego, G., Delbruck, T., Orchard, G., Bartolozzi, C., Taba, B., Censi, A., Leutenegger, S., Davison, A.J., Conradt, J., Daniilidis, K., Scaramuzza, D.: {Event-Based Vision: A Survey}. IEEE Transactions on Pattern Analysis and Machine Intelligence  \textbf{44},  154--180 (2022). \doi{10.1109/TPAMI.2020.3008413}

\bibitem{gehrig2020}
Gehrig, D., Gehrig, M., Hidalgo-Carrio, J., Scaramuzza, D.: Video to events: Recycling video datasets for event cameras. In: Proceedings of the IEEE/CVF Conference on Computer Vision and Pattern Recognition (CVPR) (June 2020)

\bibitem{heine2013}
Heine, F.: Merkel buzzed by mini-drone at campaign event. Der Spiegel  (2013), \url{https://www.spiegel.de/international/germany/merkel-campaign-event-visited-by-mini-drone-a-922495.html}

\bibitem{kirkland2019}
Kirkland, P., Di~Caterina, G., Soraghan, J., Andreopoulos, Y., Matich, G.: {UAV Detection: A STDP Trained Deep Convolutional Spiking Neural Network Retina-Neuromorphic Approach}. In: Artificial Neural Networks and Machine Learning -- ICANN 2019: Theoretical Neural Computation. pp. 724--736 (2019)

\bibitem{mozur2024}
Mozur, P., Satariano, A.: {A.I. Begins Ushering In an Age of Killer Robots}. The New York Times  (2024), \url{https://www.nytimes.com/2024/07/02/technology/ukraine-war-ai-weapons.html}

\bibitem{neftci2019}
Neftci, E.O., Mostafa, H., Zenke, F.: {Surrogate gradient learning in spiking neural networks: Bringing the power of gradient-based optimization to spiking neural networks}. IEEE Signal Processing Magazine  \textbf{36},  51--63 (2019). \doi{10.1109/msp.2019.2931595}

\bibitem{sanket2021}
Sanket, N., Singh, C., Parameshwara, C., Fermüller, C., de~Croon, G., Aloimonos, Y.: {EVPropNet: Detecting drones by finding propellers for mid-air landing and following}. In: {Robotics: Science and Systems XVII} (2021). \doi{10.15607/rss.2021.xvii.074}

\bibitem{sorbaro2020optimizing}
Sorbaro, M., Liu, Q., Bortone, M., Sheik, S.: Optimizing the energy consumption of spiking neural networks for neuromorphic applications. Frontiers in neuroscience  \textbf{14},  516916 (2020)

\bibitem{sovrasov2018}
Sovrasov, V.: ptflops: a flops counting tool for neural networks in pytorch framework (2018-2023), \url{https://github.com/sovrasov/flops-counter.pytorch}

\bibitem{stewart2021}
Stewart, T., Drouin, M.A., Gagne, G., Godin, G.: Drone virtual fence using a neuromorphic camera. In: International Conference on Neuromorphic Systems 2021 (2021). \doi{10.1145/3477145.3477264}

\bibitem{stewart2022}
Stewart, T., Drouin, M.A., Picard, M., Djupkep~Dizeu, F.B., Orth, A., Gagn\'{e}, G.: A virtual fence for drones: Efficiently detecting propeller blades with a dvxplorer event camera. In: Proceedings of the International Conference on Neuromorphic Systems 2022 (2022). \doi{10.1145/3546790.3546800}

\bibitem{synsense-speck-dev-kit-datasheet}
{SynSense}: Speck Dev Kit August Datasheet (2023), \url{https://www.synsense.ai/wp-content/uploads/2023/08/Speck-Dev-Kit-Datasheet.pdf}, accessed: 2023-12-20

\bibitem{synsense-sinabs}
SynSense: Sinabs (2024), \url{https://sinabs.ai}

\bibitem{synsense-dynapcnndoc}
{SynSense}: sinabs-dynapcnn documentation (2024), \url{https://synsense.gitlab.io/sinabs-dynapcnn/index.html}, accessed: 2024-04-12

\bibitem{wang2021}
Wang, J., Liu, Y., Song, H.: Counter-unmanned aircraft system(s) (c-uas): State of the art, challenges, and future trends. IEEE Aerospace and Electronic Systems Magazine  \textbf{36}(3),  4--29 (2021). \doi{10.1109/MAES.2020.3015537}

\bibitem{yao2024}
Yao, M., Richter, O., Zhao, G., Qiao, N., Xing, Y., Wang, D., Hu, T., Fang, W., Demirci, T., De~Marchi, M., Deng, L., Yan, T., Nielsen, C., Sheik, S., Wu, C., Tian, Y., Xu, B., Li, G.: {Spike-based dynamic computing with asynchronous sensing-computing neuromorphic chip}. Nature communications  \textbf{15}, ~4464 (2024). \doi{10.1038/s41467-024-47811-6}

\bibitem{ziegler2024spiking}
Ziegler, A., Vetter, K., Gossard, T., Tebbe, J., Zell, A.: Spiking neural networks for fast-moving object detection on neuromorphic hardware devices using an event-based camera. arXiv preprint arXiv:2403.10677  (2024)

\end{thebibliography}
\end{document}